\documentclass[10pt, conference]{IEEEtran}
\usepackage{cite}
\usepackage{amsmath,amssymb,amsfonts}
\usepackage{amsthm}
\usepackage{tabularx}
\usepackage[ruled]{algorithm2e}
\usepackage{algpseudocode}
\usepackage{graphicx}
\usepackage{textcomp}
\usepackage{xcolor}
\usepackage{subcaption}
\usepackage{multirow}
\usepackage{booktabs}
\usepackage{makecell}
\usepackage{bbm}
\usepackage{dblfloatfix}
\usepackage{threeparttable}
\usepackage{makecell}

\title{Retrieval Augmented Anomaly Detection (RAAD): Nimble Model Adjustment Without Retraining}

\author{
    \begin{tabular}{c c c c}
        \makecell{\small{Sam Pastoriza} \\ \textit{\small{Deloitte \& Touche LLP}} \ \\ \small{New York, USA} \\ \small{spastoriza@deloitte.com}} \hspace{7mm} & 
        \makecell{\small{Iman Yousfi} \\ \textit{\small{Deloitte \& Touche LLP}} \\ \small{New York, USA} \\ \small{iyousfi@deloitte.com}} \hspace{7mm} & 
        \makecell{\small{Christopher Redino} \\ \textit{\small{Deloitte \& Touche LLP}} \\ \small{New York, USA} \\ \small{credino@deloitte.com}} \vspace{4mm} &
        \makecell{\small{Marc Vucovich} \\ \small{USA} \\ \small{mdvucovich@gmail.com}} \hspace{7mm} \\
        \makecell{\small{Abdul Rahman} \\ \textit{\small{Deloitte \& Touche LLP}} \\ \small{New York, USA} \\ \small{abdulrahman@deloitte.com}} \hspace{7mm} & 
        \makecell{\small{Sal Aguinaga} \\ \textit{\small{Deloitte \& Touche LLP}} \\ \small{Indiana, USA} \\ \small{saaguinaga@deloitte.com}} \hspace{7mm} &
        \makecell{\small{Dhruv Nandakumar} \\ \small{USA} \\ \small{dhruv.nandakumar@icloud.com}}  
        \vspace{-4mm}
    \end{tabular}
}

\begin{document}

\IEEEoverridecommandlockouts
\IEEEpubid{\makebox[\columnwidth]{979-8-3315-0993-4/25/\$31.00 \copyright 2025 IEEE \hfill} \hspace{\columnsep}\makebox[\columnwidth]{ }}

\maketitle

\begin{abstract}
We propose a novel mechanism for real-time (human-in-the-loop) feedback focused on false positive reduction to enhance anomaly detection models. It was designed for the lightweight deployment of a behavioral network anomaly detection model. This methodology is easily integrable to similar domains that require a premium on throughput while maintaining high precision. In this paper, we introduce Retrieval Augmented Anomaly Detection, a novel method taking inspiration from Retrieval Augmented Generation. Human annotated examples are sent to a vector store, which can modify model outputs on the very next processed batch for model inference. To demonstrate the generalization of this technique, we benchmarked several different model architectures and multiple data modalities, including images, text, and graph-based data. 
\end{abstract}


\section{Introduction}

Cybersecurity artificial intelligence (AI) models designed for network intrusion threat detection require very high, but nuanced, model precision. False positive (FP) reduction is crucial for the practical implementation of anomalous behavior detection. Compared to false negatives (FN), where threats go undetected, even a small FP rate can render a strong model impractical and negatively impact business operations. Enterprise networks monitor their network traffic using high-speed monitoring systems using protocols such as NetFlow and IPFIX \cite{6814316}. During the cyber threat detection model inference process, even a sub-percent FP rate can cause alert fatigue for the security analyst, resulting in mismanaged response capacity. Maintaining a high precision for the anomaly detection model requires significant computational resources and training time. 

During model development, there is a trade-off between training on only a target network or multiple networks. In a single network, the model is at risk of mistakenly learning a threat actor's behaviors as normal if they are already acting discretely on the network. Models trained on other networks have better generalization abilities, but likely still require further tuning to the particular target network. A reasonable solution seems to be tuning these generalizable models only with high-quality, vetted data from the target network, which often means hand labeling. These vetted, hand labeled examples are difficult to collect in large quantities quickly. It is also difficult to course correct model performance on just a few examples like allow lists, block lists, and other simple rule-based approaches. These are counter-intuitive to the motivation of using flexible deep learning-based methodologies in the first place, and often make the precision worse, not better. 

This paper introduces a novel method to efficiently provide feedback to models without retraining. Retrieval Augmented Generation (RAG) \cite{lewis2021retrievalaugmentedgenerationknowledgeintensivenlp} has empowered large language models (LLMs) to become more powerful without retraining. An input can be “augmented” with additional information to enable the underlying LLM to produce more accurate responses without storing that information in its underlying weights. Instead of augmenting the inputs to a model, we propose an approach that alters the outputs of a model to prevent similar mistakes from being made. We call this approach Retrieval Augmented Anomaly Detection (RAAD). RAAD allows for real-time human feedback, which is a powerful tool that allows a user to make corrections as needed. By identifying mistakes and storing them, a pipeline can check for similar mistakes and adjust its predictions based on similarity metrics. This approach also simplifies the collection of data for retraining in the future, as inputs are human reviewed and annotated. In general, the key contributions of this paper are as follows.

\begin{itemize}
    \item A novel and generalizable method of introducing real time feedback into a model’s pipeline
    \item A technique to apply this to anomaly detection based systems
    \item Benchmarking performance of the proposed method on three modalities of data: image, text, and graph
\end{itemize}

\IEEEpubidadjcol


The paper begins with a review of the literature discussing previous work in the field of LLMs and various semi-supervised learning techniques, followed by a description of the training methodology, implementation details, and results. We then conclude with remarks on some of our key design decisions and a description of possible future work.

\section{Related Work} \label{sec:related_work}

The reduction of FP rates is a common challenge across machine learning applications and techniques, from Generative AI (GenAI) techniques like RAG to modeling methodologies like deep semi-supervised learning. Outdated information or inaccuracies in LLM training data have led to RAG methodologies\cite{lewis2020retrieval}. RAG allows LLMs to incorporate additional data sources before making predictions based on user input \cite{shuster2021retrieval}. However, the fundamental core of RAG has been shown to have promising results with a variety of model types along with LLMs.  Pan et al. used RAG to help facilitate cyber investigations with system logs \cite{PanLogs}. Their architecture uses an LLM to perform semantic analysis between log samples retrieved from the vector database and the queried log entry. The logs stored in the vector database are the vector embedding of known normal logs. Therefore, when the retrieval score matches the criteria, such as the highest similarity score or the minimum threshold score, the vector database returns the resultant embedding vectors \cite{PanLogs}. Although their results with a cyber-focused RAG model are promising, LLMs are not always usable in cybersecurity due to the variety and sensitivity of data types.  Other authors, such as Al Jallad et al. propose using larger amounts of data to help train generalizable deep learning model to detect anomalies \cite{Al_Jallad_2020} with lower false positive rates in a vein similar to the anomaly detection model for which RAAD was originally developed, although we specifically follow the architecture of Nandakumar et al. \cite{nandakumar2023foundationalmodelsmalwareembeddings}.

Deep semi-supervised learning techniques have evolved over time to improve model performance and reduce labeling costs. Ouali et al. defines the goal of semi-supervised learning as leveraging the unlabeled data to produce a prediction function with trainable parameters \cite{ouali2020overview}. Lee \cite{Lee2013PseudoLabelT} introduced the concept of pseudo-labeling, which involves generating proxy labels to augment the training set. This was further expanded by combining label propagation with pseudo-labeling in Iscen et al. \cite{DBLP:journals/corr/abs-1904-04717}, labeled sample constraints in Arazo et al\cite{DBLP:journals/corr/abs-1908-02983}, and retraining models with regularization and pseudo-labeling in Sohn et al \cite{DBLP:journals/corr/abs-2001-07685}. Although effective in reducing FP rates, it is not a long term sustainable method for an anomaly detection model in a robust production environment, where speed and efficiency are priority. These methods often require significant computational resources and training time. RAAD does not require constant retraining and provides an alternative that is low-touch in deployment. The architectures reviewed here inform our approach and methodology towards reducing false positives in an anomaly detector. 

\section{Methodology} \label{sec:methodology}

In this section, we will cover the datasets we used to test our RAAD approach, as well as the architecture we developed and how it is implemented to adjust probability- and loss-based predictive models.

\small
    \begin{table*}[t]
        \centering \renewcommand{\arraystretch}{1.4}
        \caption{RAAD Datasets}
    \begin{tabularx}{\textwidth}{|l|X|X|}
    \hline
    \bfseries{\thead{Dataset}} & \bfseries{\thead{Description}} & \bfseries{\thead{Notes}} \\
    \hline
    MNIST & 60,000 training examples and 10,000 test examples of handwritten numeric digits in black and white &  Widely used to train image recognition models. No preprocessing was performed outside of a secondary 80-20 train/validation split. \\
    \hline
    E-MNIST & 697,932 training examples and 116,323 test examples of 62 different classes of handwritten characters in black and white & There are six datasets, varying by type of data, including just letters, just digits and different mixes of the two. We used the default ``by-class" dataset, which is unbalanced and includes upper and lowercase letters and digits. No preprocessing was performed outside of a secondary 80-20 train/validation split.  \\
    \hline
    Malicious URLs & A set of 640,000+ urls combining several datasets, including ISCX-IRL-2016, faizen, Phishtank and Phishstorm, among others.  & Labels in this dataset include benign, defacement, phishing and malware. Preprocessing included removing special characters, splitting the data, then tokenizing and training a FastText embedding model. \\
    \hline
    NC-CDC & Consists of labeled blue-team and red-team data from real attack simulations on a cyber range.  & Data spanned two consecutive days in 2020 and in 2021. Attack types consist of Network Scanning, Interrogation, Botnet, and Command and Control. Data was preprocessed in the same manner as \cite{nandakumar2023foundationalmodelsmalwareembeddings}.  \\
    \hline
    MAWI & Real network PCAP data from hundreds of devices across 12 universities in Japan. & Flow data for 7 consecutive days from 2021, and 1 days from 2016 were used from a dataset of over 14 years. Dataset is not labeled but is considered benign due to extreme imbalance in real-world networks. Data was preprocessed in the same manner as \cite{nandakumar2023foundationalmodelsmalwareembeddings}.  \\
    \hline
    \end{tabularx}
    \label{table:datasets}
    \end{table*}
\medskip

\normalsize

\subsection{Datasets}

The data sets used to test our approach span several modalities, including image data sets such as MNIST (Modified National Institute of Standards and Technology) \cite{lecun2010mnist} and E-MNIST (Extended MNIST) \cite{cohen_afshar_tapson_schaik_2017}, text data sets such as the malicious URL (Uniform Resource Locator) dataset \cite{maliciousurls}, and graph-based datasets, such as NetFlow \cite{li2013survey} connections used in our zero-day threat detection models \cite{nandakumar2023foundationalmodelsmalwareembeddings}. Table~\ref{table:datasets} describes these datasets in more detail.

\subsection{RAAD Architecture} \label{sec:architecture}

\begin{figure}
    \centering
    \captionsetup{singlelinecheck=false, format=hang, justification=raggedright, font=footnotesize, labelsep=period}
    \begin{subfigure}[b]{0.44\textwidth}
        \includegraphics[width=\textwidth]{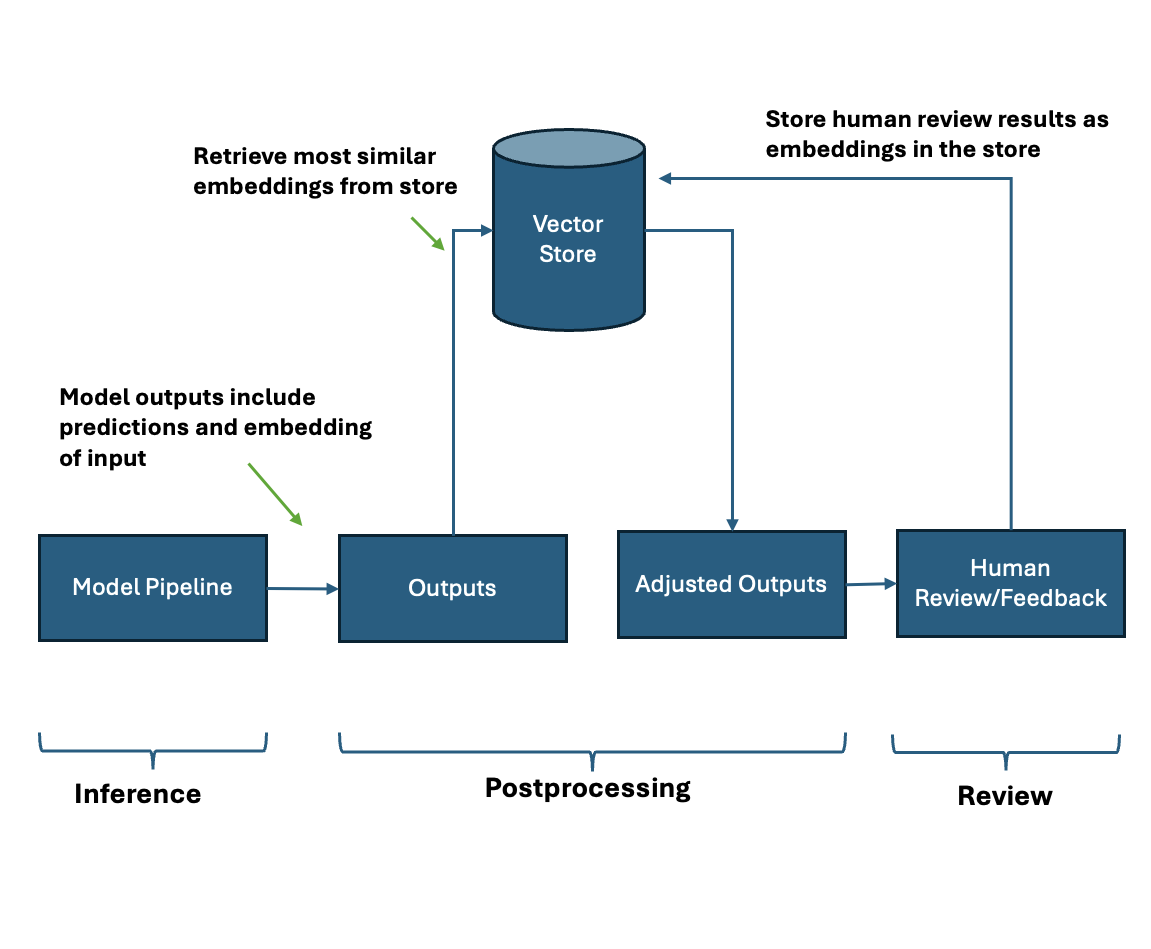}
    \end{subfigure}
    \caption{RAAD Architecture}
    \label{fig:raad_architecture}
\end{figure}

Traditional ML architectures generally include raw input data collection, pre-processing, feature engineering, model inference, post-processing, and review. Our RAAD architecture (Fig.~\ref{fig:raad_architecture}), consists of a traditional machine learning (ML) architecture, with the addition of a post-processing step to adjust model outputs. It allows for direct modification of model outputs based on human feedback in real time. As part of the pipeline, the model can be configured to output both a prediction and a representation of the input as a learned embedding. As humans provide feedback on the quality of the model, they mark when models get predictions incorrect, and those incorrect predictions are stored in a permanent location as a learned embedding. Once that embedding is stored, the next batch of model outputs can utilize that knowledge by comparing its own learned embeddings to those stored. Based on the similarity scores generated by the input embeddings, each of the outputs can be adjusted to prevent past mistakes by the model. Depending on the type of model, the model outputs are adjusted approach using RAAD as described in the subsequent sections.

\subsection{Probability Bounded Adjustments}

Many models output probabilities, typically produced by an activation function such as Softmax. To adjust probabilities (and losses), we must be careful to adjust results based on the knowledge of the performance of the existing model and the environment in which the model operates. To accommodate a reasonable amount of flexibility, we introduce three hyperparameters that help adjust the probabilities as described in Algorithm \ref{alg:probability_bounded_adjustements}. This method takes into account the similarity and distance between the inputs and known false positives and adjusts the probabilities so inputs closer to false positives have their respective probabilities dropped below a defined threshold, thus changing the overall prediction of the model. Additionally, hyperparameters do not require much adjustment across models, as seen in Table \ref{table:dataset_results}. 

\begin{algorithm}[t]
\caption{Probability Bounded Adjustments}
\label{alg:probability_bounded_adjustements}
\begin{algorithmic}[1]
\State 
\textbf{Input:} 
\begin{itemize}
    \item $\mathcal{V_{\text{init}}}$: Embedding Representation
    \item $\mathcal{P_{\text{init}}}$: Initial Probabilities
    \item $\mathcal{\mathcal{V_{\text{fp}}}}$: Annotated False Positives Embeddings
\end{itemize}

\State
\textbf{Hyperparameters:} 
\begin{itemize}
    \item $ \tau \in (0,1)$: Cosine similarity threshold
    \item $ \alpha \in (10, 20, ..., 100)$: Sharpness of fitted polynomial
    \item $ \delta \in \mathbb{N}$: Optional euclidean distance threshold
\end{itemize}

\State 
\textbf{Find Similar Vectors}
\begin{enumerate}
    \itemsep 0.5em 
    \item $\mathcal{\theta_{\text{all}}} \leftarrow \frac{v_{\text{init}}^{(i)} \cdot v}{\|v_{\text{init}}^{(i)}\|\|v\|}, \quad \forall v_{\text{init}}^{(i)} \in \mathcal{V}_{\text{init}}, v \in \mathcal{V}_{\text{fp}}$
    \item $\mathcal{\theta_{\text{closest}}} \leftarrow \max{\mathcal{\theta_{\text{all}}}}$
    \item $\mathcal{V_{\text{closest}}} \leftarrow \arg\max{\mathcal{\theta_{\text{all}}}}$
    \item $d_{\text{closest}} \leftarrow \sqrt{\sum_{i=1}^{n} (\mathcal{V_{\text{fp}_\text{i}}} - \mathcal{V_{\text{closest}_\text{i}}})^2}$
\end{enumerate}

\State
\textbf{Fit Polynomial Adjustment Function}
\begin{enumerate}
    \itemsep 0.5em 
    \item $f(\theta) = LeastSquaresFit(P, D)$
    \begin{itemize}
        \item $P \in \{(0,0), (\tau, \tau)\}$: Data points on curve
        \item $D = \alpha$: Degree of polynomial
    \end{itemize}
\end{enumerate}

\State
\textbf{Adjust Similarity Score and Distances}
\begin{enumerate}
    \itemsep 0.5em 
    \item $\mathcal{\theta_{\text{adjusted}}} = \begin{cases} \mathcal{\theta_{\text{closest}}} & \text{if } \mathcal{\theta_{\text{closest}}} \geq \tau, \\ f(\mathcal{\theta_{\text{closest}}}) & \text{otherwise}. \end{cases}$
    \item $d_{\text{adjusted}} = \begin{cases} 1 & \text{if } \delta = \emptyset\, \\ min(\frac{\delta}{d_{\text{closest}}}, 1) & \text{otherwise}.  \end{cases}$
\end{enumerate}

\State
\textbf{Adjust Probability}
\begin{enumerate}
    \itemsep 0.5em 
    \item $FP_{confidence\_score} = FP_{cs} = \mathcal{\theta_{\text{adjusted}}} * d_{\text{adjusted}}$
    \item $\mathcal{P_{\text{adjusted}}} = \mathcal{P_{\text{init}}} * (1 - FP_{cs})$
\end{enumerate}

\State \textbf{Output:} 
$\mathcal{P_{\text{adjusted}}}$ $\rightarrow$ \text{Adjusted Probabilities}

\end{algorithmic}
\end{algorithm}

\subsection{Loss Bounded Adjustments}

Some models just output a loss directly, which requires an additional step when making adjustments using RAAD, as these outputs are unbounded, unlike probabilities. A sigmoid curve is modified to account for this infinite upper bound. Specifically, instead of an adjusted similarity score, the output of a sigmoid curve for loss-based adjustments is treated as more of a multiplicative/fractional adjustment factor, where similar embeddings with scores closer to 1 would be adjusted by multiplying a factor closer to 0. In a similar method to Algorithm \ref{alg:probability_bounded_adjustements}, Algorithm \ref{alg:losses_bounded_adjustements} describes how these losses are adjusted so that inputs that are close to false positives have their respective losses dropped below a threshold, thus changing the overall prediction.

\begin{algorithm}[t]
\caption{Loss Bounded Adjustments}
\label{alg:losses_bounded_adjustements}
\begin{algorithmic}[1]
\State 
\textbf{Input:}
\begin{itemize}
    \item $\mathcal{V_{\text{init}}}$: Embedding Representation
    \item $\mathcal{L_{\text{init}}}$: Initial Losses
    \item $\mathcal{\mathcal{V_{\text{fp}}}}$: Annotated False Positives Embeddings
\end{itemize}

\State
\textbf{Hyperparameters:} 
\begin{itemize}
    \item $ \tau \in (0,1)$: Cosine similarity threshold
    \item $ \alpha \in (10, 20, ..., 100)$: Sharpness of fitted polynomial
    \item $ \delta \in \mathbb{N}$: Optional euclidean distance threshold
\end{itemize}

\State 
\textbf{Perform Algorithm \ref{alg:probability_bounded_adjustements}}

\State
\textbf{Adjust Losses}

$\mathcal{L_{\text{adjusted}}} = \mathcal{L_{\text{init}}} * (\cfrac{-1}{1 + e^{\alpha * (\tau - FP_{cs})}} + 1)$

\State \textbf{Output:} $\mathcal{L_{\text{adjusted}}}$ $\rightarrow$ \text{Adjusted Losses}

\end{algorithmic}
\end{algorithm}

\section{Experimental Design} \label{sec:design}

Due to the natural imbalance of these datasets, and the highly skewed tendency towards negative, or benign events, we utilized Precision, Recall, and the area under the Receiver Operating Characteristics (ROC) curve (AUC) to evaluate our models. The goal when working with each of these datasets was to create and fit several models per dataset, evaluate them, assess the outputs, and add failed predictions to a store of learned embeddings. These annotated failed predictions are typically false positives. Then we rerun the pipeline, adjust the sharpness and similarity threshold values, and re-evaluate the adjusted results.

\subsection{Design of Embeddings}

The RAAD mechanism works better with models that have an embedding that accurately represents the data so that similar data points are represented similarly in the learned embedding space. Ideally, the model RAAD is applied to should follow the cluster assumption\cite{ouali2020overview}, if points are in the same cluster, they are likely to be of the same class \cite{4787647}. For neural networks, this can be as simple as choosing a layer from the neural network that is large enough to store differentiations in the data. A simple example of this would be choosing the bottleneck layer in an autoencoder, as this layer is the smallest layer that contains the most important information used to reconstruct the input. Typically, several different layers of a neural network are tried before determining the most effective layer. In assessing whether an embedding space for a model is sufficiently separated for a successful application of RAAD a good test is to simply leverage the Jaccard index \cite{JaccardDistributionDL} of the embedding space, treating distinct classes as their own sets for comparison. 

\begin{figure}
    \centering
    \captionsetup{singlelinecheck=false, format=hang, justification=raggedright, font=footnotesize, labelsep=period}
    \begin{subfigure}[b]{0.45\textwidth}
        \includegraphics[width=\textwidth]{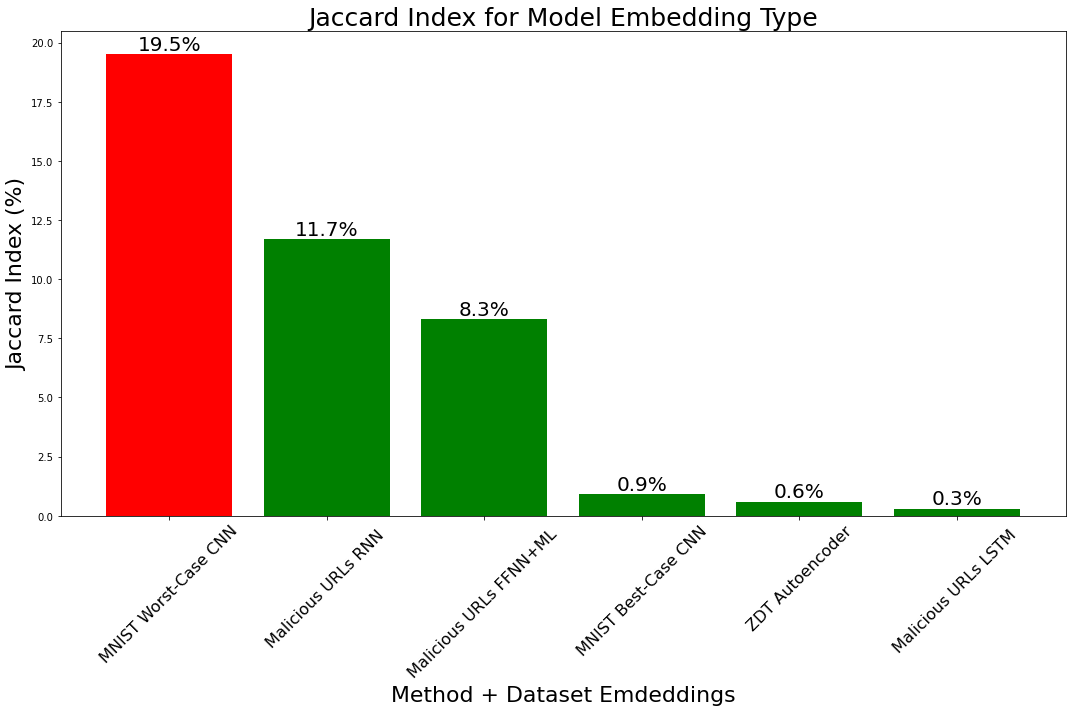}
        
    \end{subfigure}
    \caption{Jaccard Index of several models and embeddings}
    \label{fig:jaccard_index}
\end{figure}

A Jaccard index around or below $10\%$, has been shown to be a good indicator that a RAAD implementation would be of use. This nuanced application of the Jaccard index focuses on the dissimilarity of the embedding space. A low Jaccard index signifies minimal overlap between the sets of data points classified by each model. This reflects distinct decision boundaries, which is desirable when predicting the success of RAAD. By having a clear separation, it enhances the model's ability to capture specific features, making it a strong indicator for success.

Embeddings for RAAD need not come from a deep learning model, and can also come from embedding algorithms such as Word2Vec \cite{mikolov2013efficientestimationwordrepresentations}, GloVe \cite{pennington-etal-2014-glove}, or FastText \cite{joulin2016bagtricksefficienttext} for text.  Likewise, for graph data sets, an algorithmic embedder such as FastRP \cite{chen2019fastaccuratenetworkembeddings}, can be used. \cite{nandakumar2023foundationalmodelsmalwareembeddings}.

\subsection{Tuning for Sharpness and Similarity Threshold}

Choosing the hyperparameters is critical to the success of RAAD. The sharpness value can be thought of as the steepness of the dividing plane between a range of similarities, while the similarity threshold is the upper bound of the similarity scores. A sharpness closer to 0 is more flexible, but may allow for missed true positives, while a sharpness closer to 100 is more rigid, reducing only the most confident of false positives. The similarity threshold is the similarity score at which we are most confident the learned embedding is a false positive based on the proximity to an embedding in the database.

\begin{figure*}
    \centering
    \captionsetup{singlelinecheck = false, format= hang, justification=raggedright, font=footnotesize, labelsep=period}
    \begin{subfigure}[b]{0.43\textwidth}
        \includegraphics[width=\textwidth]{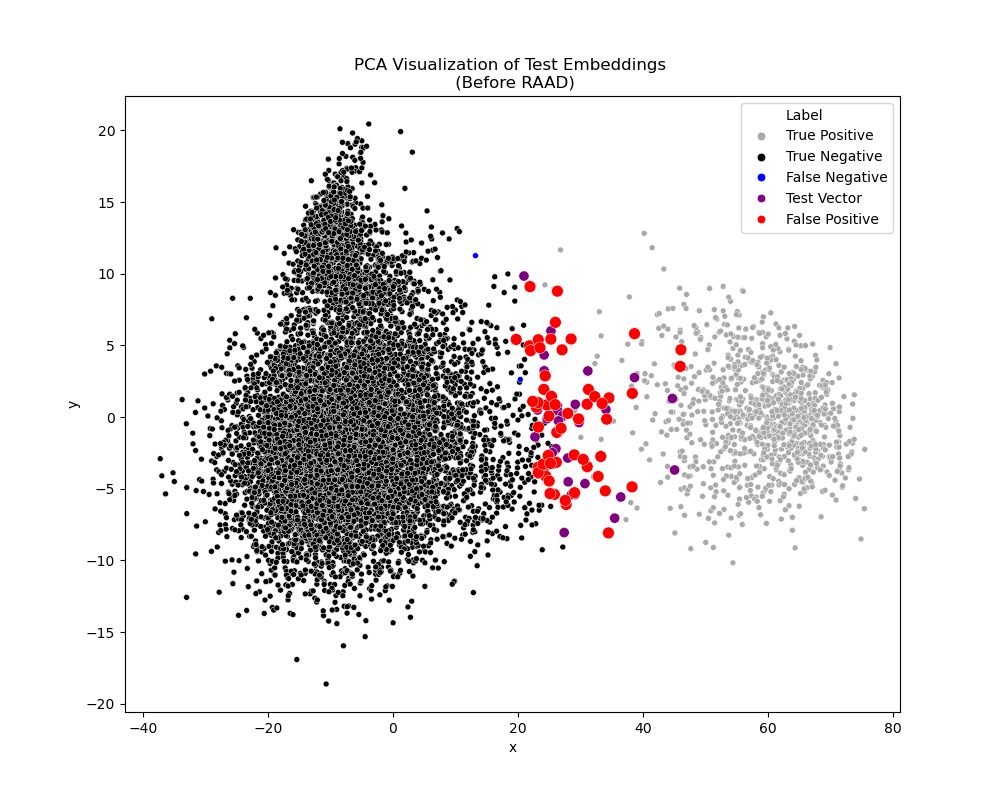}
    \end{subfigure}
    \begin{subfigure}[b]{0.43\textwidth}
        \includegraphics[width=\textwidth]{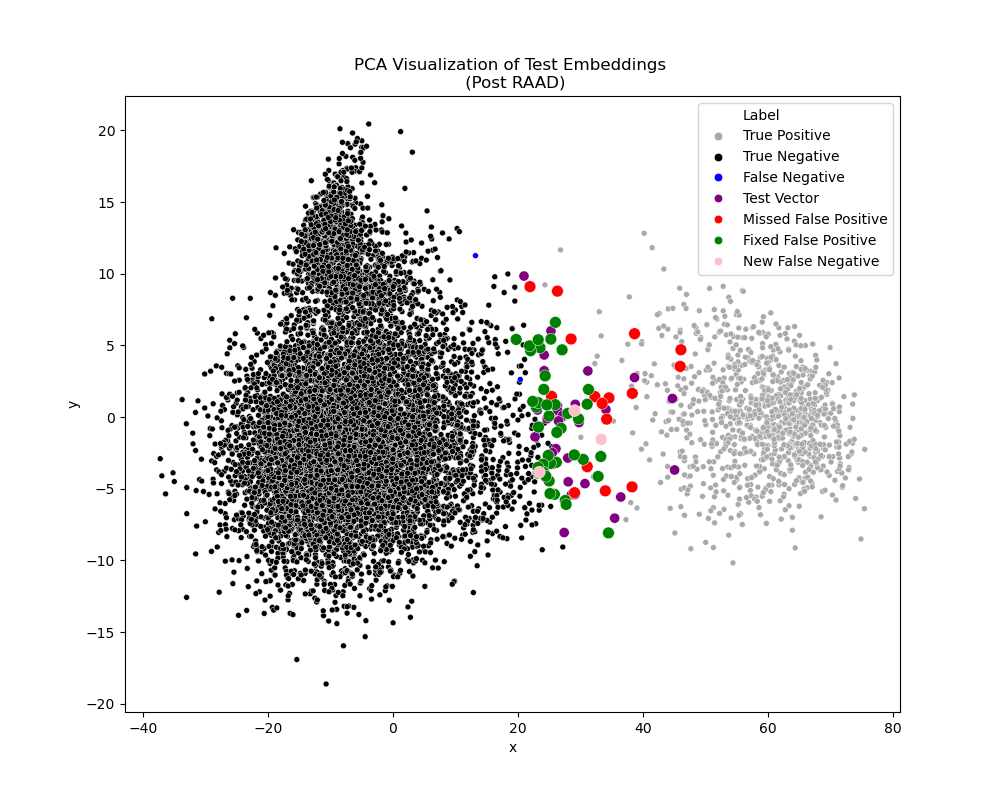}
    \end{subfigure}
    \caption{MNIST Example (Before/After RAAD)}
    \label{fig:minst_raad}
\end{figure*}

\section{Results and Discussion} \label{sec:results_and_discussion}

\begin{table*}[b!]
\centering
\scriptsize\renewcommand{\arraystretch}{1.4}
\caption{RAAD Results by Dataset}
\begin{tabular}{|p{1.75cm}|p{1.3cm}|p{1.75cm}|p{1.1cm}|p{1.5cm}|p{1cm}|p{1cm}|p{1cm}|p{1cm}|p{1cm}|p{1cm}|}
\hline

\bfseries{Dataset} & \bfseries{Model} & \bfseries{Embedding Size} & \bfseries{Sharpness} & \bfseries{FP Threshold} & \bfseries{Distance} & \bfseries{$\bf{F1_{\text{orig}}}$\textsuperscript{a}} & \bfseries{$\bf{F1_{\text{new}}}$\textsuperscript{a}} & \bfseries{$\triangle$FPs\textsuperscript{a}} & \bfseries{$\triangle$TPs\textsuperscript{a}}& \bfseries{$\bf{FPs_{\text{orig}}}$\textsuperscript{a}} \\
\hline
MNIST & CNN & 256 & 60 & 0.95 & (3-5) & 98.08\% & 98.58\% & -58\textsuperscript{b} & -14\textsuperscript{b} & 58 \\
\hline
Malicious URLs & FFNN+ML & 128 & 60 & 0.95 & 1 & 93.76\% & 93.79\% & -36 & -7 & 1433\\
Malicious URLs & RNN & 128 & 60 & 0.95 & 1 & 92.96\% & 93.08\% & -121 & -32 & 1291 \\
Malicious URLs & LSTM & 128 & 60 & 0.95 & 1 & 93.69\% & 93.79\% & -116 & -40 & 1497 \\
\hline
ZDT & Autoencoder & 6 & 70 & 0.95 & 1 & 95.4\% & 96.4\% & -9185 & 0 & 9200 \\
\hline
\end{tabular}
\label{table:dataset_results}
\footnotesize \\
\raggedleft{\textsuperscript{a} FP = False Positive, TP = True Positive, F1 = F1 Score}
\\
\footnotesize
\raggedleft{\textsuperscript{b} The best result after fitting a one-versus-all CNN on MNIST. Results for other classes average at a 39\% reduction in false positives}
\end{table*}

RAAD can be broadly applicable across modalities where false positives happen. In addition to fitting RAAD to an anomaly detection dataset, we also show it works with image data and text data datasets. 

\subsection{Graph/Anomaly Detection Dataset Results} \label{sec:graph_dataset_results}

RAAD was initially created to improve our zero day threat detection model. While training this model, primarily on network flow traffic data, the training sets are never totally comprehensive, meaning not every "normal" network connection is tagged as normal. During the deployment of this model, we found it makes consistent mistakes that are particular to the network it is operating on. This happens even in deployments with a precision over 99\% \cite{nandakumar2023foundationalmodelsmalwareembeddings}. By using RAAD, the embeddings associated with these false positives are stored and future events with a similar embedding would be tagged as a false positive and not considered an anomaly.

To demonstrate how RAAD works with this model, we fit an autoencoder to several network traffic datasets as described in Table \ref{table:datasets}. We undertrained the model slightly so the model did not learn all of the behaviors of the training dataset. Using RAAD, we found a sharpness of \textbf{70}, a false positive threshold of \textbf{0.98}, and a max distance of \textbf{1} performed best, as seen in Table \ref{table:dataset_results}. We reduced the number of false positives from an original value of \textbf{9200}, down to just \textbf{15}, while not introducing \textbf{any} new false negatives. This is strong evidence this approach was applied successfully.

\subsection{Image Dataset Results}

To apply RAAD to the MNIST and E-MNIST, we needed to convert the dataset from a multiclass classification problem to a binary classification problem. To do this, we trained a one-versus-all classification model per class and checked to see that RAAD worked. To be consistent, we used an embedding size of 256 all for MNIST, and 512 for E-MNIST. The difference in embedding size can be attributed to the size of the dataset and increase in number of classes. The results of applying RAAD for MNIST and E-MNIST can be found in Table \ref{table:dataset_results}. In general, a sharpness of \textbf{60}, threshold of \textbf{0.95} and max distance between \textbf{3-5} had the best false positive reduction rates observed. In addition to these promising results, we can visualize a successful application of RAAD and show how the space between the clusters of positive and negative classes can be better separated, as seen in Figure \ref{fig:minst_raad}.

\subsection{Text Dataset Results}

 To show the viability of RAAD, several models were fit to this dataset. When fitting a recurrent neural network on this dataset, we found a sharpness of \textbf{60}, a threshold of \textbf{0.95}, and a max distance parameter of \textbf{1} performed best. The Long Short-Term Memory (LSTM) performed slightly better, but similar parameters were used when applying RAAD. Since the LSTM performed generally better, as seen in Table \ref{table:dataset_results}, the sharpness and threshold were both higher, indicating a better separation of the embedding space. This shows that RAAD has the flexibility with a variety of parameters users can tune while also allowing for strong performance while using only the default parameters.

\section{Conclusions, Limitations, and Future Work}

In this work, we have presented a novel method of providing feedback to models without retraining by comparing identified mistakes made by models to future outputs. In the proposed methodology, we utilize underlying embeddings of inputs that were mistakenly identified as points of comparison when new inputs are fed into the model. By utilizing a vector database, in similar fashion to the RAG methodology, we can store those mistakes and use them during post-processing to augment model outputs and catch similar mistakes. This approach allows for real-time human feedback that can affect model output immediately after a mistake is identified. Additionally, this can be used as a store of well labeled data for retraining the model when needed, as humans are reviewing and labeling these data points. With enough well-labeled data points, retraining will improve the model's performance, but as this solution is meant for real-time feedback, this approach serves as an excellent option before needing to retrain. We believe our results show strong indicators this method has broad applicability across different modalities and types of models and embeddings. 

RAAD has shown especially strong performance in certain cases based on our metrics. Generally, RAAD can make a highly precise model even better, but it is unlikely to improve a poor or even good model. To capture the embedding space separability, we found that a Jaccard Index below 10\% is a good indicator. Further investigation found that models that were able to perform at a high level made fewer mistakes, and the mistakes they made often were consistent. This meant the embeddings were well separated and consistent in nature. A model that had never seen a certain input might make a mistake, but the underlying embedding of that input would be separated from other examples, so those mistakes are easily isolated and corrected. Models that performed poorly did not have an embedding space that differentiated inputs enough such that RAAD could isolate and correct those problems. 

As the quality of embedding spaces are key to RAAD's success, utilizing metric learning to separate out those embedding spaces could lead to a more generalizable application of this idea across models and modalities. A limitation of RAAD is the solution works best when the internal embedding space achieves the metrics mentioned in Figure \ref{fig:jaccard_index}. Metric learning, as originally described by Hoffer and Ailon \cite{hoffer2018deepmetriclearningusing}, aims to create a better embedding space by optimizing a function such as triplet or contrastive loss, where the distance in the embedding space preserves object similarity. This means that similar objects get closer together while distant objects are further away within the embedding space. By utilizing metric learning to augment the vector space, it could lead to significant performance improvement, both when using RAAD, but also in the model itself. It would be interesting to see if the performance of metrics of RAAD could somehow be directly incorporated into some new loss function to better guarantee that a model could be tuned in this manner.

Another path for future work in RAAD is utilization of potentially richer human feedback. For example, as it stands, this approach only takes into account binary identifiers (false positive or true positive). Instead of constraining RAAD, we could allow for additional metadata attached to these embeddings, where certain properties could have different effects that reduce or increase the probability outputted by the model. For example, certain labeled points could be more severely punished versus others that could considered less important. This kind of information could be stored as metadata in the vector database alongside the embedding. In addition to adding flexibility with varying adjustment factors based on human feedback, this solution could be further expanded to include multi-class classification scenarios. The closer/further an embedding is to a human labeled class, the more or less the probability of the embedding being related to that class could be adjusted. Utilizing more of the potential feedback a human provides could empower this framework going forward.
\bibliographystyle{IEEEtran}
\bibliography{ref}

\end{document}